\documentclass{article}

\usepackage{microtype}
\usepackage{graphicx}
\usepackage{subfigure}
\usepackage{booktabs} 

\usepackage{hyperref}

\usepackage[accepted]{icml2020}

\usepackage{amsmath, amsfonts, amsthm, amssymb}
\usepackage{enumitem}

\DeclareMathOperator*{\argmin}{arg\,min}

\icmltitlerunning{Unlabelled Data Improves Bayesian Uncertainty Calibration under Covariate Shift}

\begin{document}

\twocolumn[
\icmltitle{Unlabelled Data Improves Bayesian Uncertainty Calibration \\ under Covariate Shift}



\icmlsetsymbol{equal}{*}

\begin{icmlauthorlist}
\icmlauthor{Alex J. Chan}{cam}
\icmlauthor{Ahmed M. Alaa}{UCLA}
\icmlauthor{Zhaozhi Qian}{cam}
\icmlauthor{Mihaela van der Schaar}{cam,UCLA,turing}

\end{icmlauthorlist}

\icmlaffiliation{cam}{University of Cambridge, Cambridge, UK}
\icmlaffiliation{UCLA}{University of California, Los Angeles, USA}
\icmlaffiliation{turing}{Alan Turing Institute, London, UK}

\icmlcorrespondingauthor{Alex J. Chan}{ajc340@cam.ac.uk}

\icmlkeywords{Uncertainty, Dropout, Covariate Shift, Transductive, Bayesian Neural Networks}

\vskip 0.3in
]



\printAffiliationsAndNotice{}  

\begin{abstract}
Modern neural networks have proven to be powerful function approximators, providing state-of-the-art performance in a multitude of applications. They however fall short in their ability to quantify confidence in their predictions --- this is crucial in high-stakes applications that involve critical decision-making. Bayesian neural networks (BNNs) aim at solving this~problem~by~placing~a prior distribution over the network's parameters, thereby inducing a posterior distribution that encapsulates predictive uncertainty. While existing variants of BNNs based on Monte Carlo dropout produce reliable (albeit approximate) uncertainty estimates over in-distribution data, they tend to exhibit over-confidence in predictions made on target data whose feature distribution differs from the training data, i.e., the {\it covariate shift} setup. In this paper, we develop an approximate Bayesian inference scheme based on {\it posterior regularisation}, wherein unlabelled target data are used as ``pseudo-labels'' of model confidence that are used to regularise the model's loss on labelled source data. We~show~that~this~approach~significantly~improves the accuracy of uncertainty quantification on covariate-shifted data sets, with minimal modification to the underlying model architecture. We demonstrate the utility of our method in the context of transferring prognostic models of prostate cancer across globally diverse populations. 
\end{abstract}

\section{Introduction}

Modern neural networks have achieved the state of the art predictive performance in a wide variety of applications. They are especially useful in areas where a large quantity of labelled i.i.d data are available \cite{krizhevsky2012imagenet}.  However, neural networks fall short in their ability to quantify their confidence in the predictions, which leads to difficulties to apply them to mission critical domains. The immediate problem is that neural networks may issue \textit{erroneous predictions with high confidence} \cite{ovadia2019can}. These over-confident predictions would mislead rather than inform human experts' decisions and can lead to severe consequences in high-stakes applications. 

In practice, the task of quantifying predictive uncertainty is even more challenging because the training and testing data are typically not i.i.d due to the impact of exogenous factors over time or the inconsistency in data collection. This situation is known as \textit{covariate shift} \cite{shimodaira2000improving} and various research has indicated that this may cause neural networks to display unexpected behaviour \cite{ovadia2019can}. In the extreme case, they may confidently produce nonsensical predictions for out-of-distribution adversarial examples \cite{madry2017towards}.
While in this work we do not consider the scenario of a targeted adversarial attack, we would like the network to return a high uncertainty prediction if a test point falls far from any of the training data.
We motivate the need for work in the area with a concrete example; consider the case of trying to predict the mortality rate for a group of patients with prostate cancer in a country where we have no labels due to tight privacy regulations on medical data. We have access though to labelled examples from another country which we could use to train the model, the problem being that the populations of each country may differ in their distribution so a model purely trained on the labelled data may perform poorly on the unlabelled data both in terms of accuracy and uncertainty estimation. This type of problem is common in the medical setting and errors here are especially damaging since model predictions may have a direct impact on the treatment a patient receives.  

Bayesian neural networks (BNNs) \cite{neal2012bayesian} aim to solve the uncertainty quantification problem by learning neural networks via Bayesian inference, a principled way to reason under uncertainty. BNNs encapsulate the prediction uncertainty in the posterior predictive distribution, which is typically intractable and has to be approximated \cite{graves2011practical,blundell2015weight}. 
While existing approximation methods are able to produce reliable uncertainty estimates over in-distribution data, it has been shown that they tend to be over-confident under covariate shift. 

In this paper, we propose Transductive Dropout, a method leveraging information from the unlabelled target data to find a better approximation to the posterior. We make the following observation: a point being in the target data is an indication that the model should output higher uncertainty because the target distribution is not well-represented by training data due to covariate shift. Therefore, we use whether the data come from training or target set as a ``pseudo-label'' of model confidence. This naturally leads to a posterior regularisation term which we incorporate into the variational approximation objective. 
Our regulariser can be easily applied to many of the current network architectures and inference schemes --- here, we demonstrate its usefulness in Monte Carlo Dropout, showing that it much more appropriately quantifies its uncertainty under covariate shift. Empirical evaluations demonstrate that our method performs competitively compared to Bayesian and frequentist approaches in the task of prostate cancer mortality prediction across globally diverse populations.

\begin{figure}[t]
   \centering
   \includegraphics[width=3.25in]{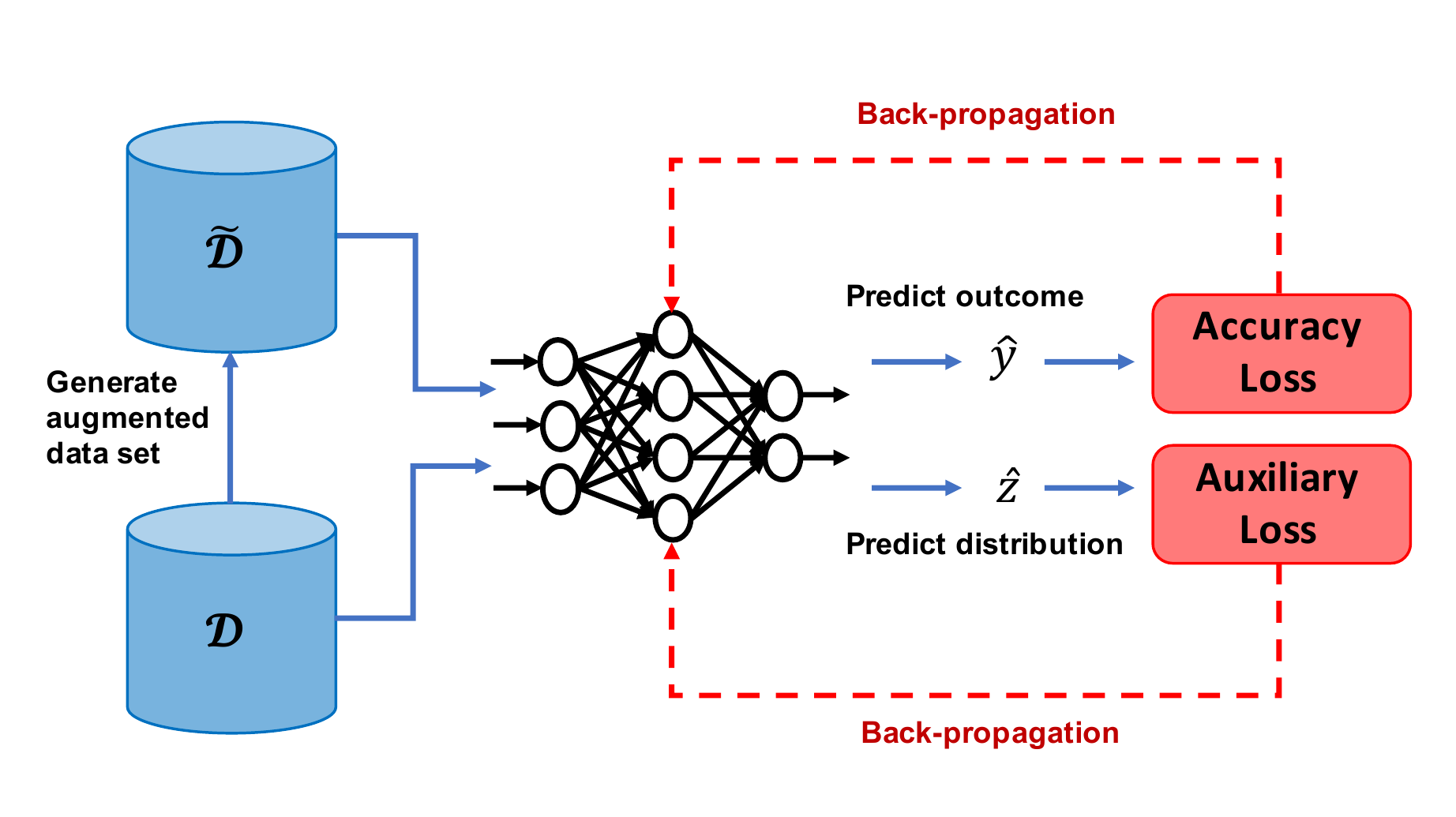}
   \vspace{-7mm}
   \caption{\footnotesize {\bf High-level depiction of our approach.} We first generate our augmented data set with pseudo-labels before feeding forward to make predictions and then back-propagating both errors through the network.} 
   \rule{\linewidth}{.75pt}  
   \label{fig:overview} 
 \vspace{-9.75mm}  
\end{figure}

\section{Related Work}

\subsection{Overview of Related Methods}
Utilising unlabelled data to improve uncertainty estimate under covariate shift is a previously less explored area in the literature. Here we highlight some of the key methods in the surrounding fields to contextualise our work.

\paragraph{Bayesian Uncertainty Estimate for Neural Networks}

Bayesian methodology has been applied to quantify the predictive uncertainty of neural networks leading to a large family of methods known as Bayesian Neural Networks (BNN). BNN learns a posterior distribution over parameters that encapsulates the model uncertainty. Due the complexity of deep neural networks, the exact posterior is usually intractable. Hence, much of the research in BNN literature is devoted to finding better approximate inference algorithms for the posterior. Popular approximate Bayesian approaches include dropout-based variational inference \citep{gal2016dropout, kingma2015variational}  and Stochastic Variational Bayesian Inference \citep{blundell2015weight,graves2011practical,louizos2017multiplicative}. These methods are known to achieve reliable uncertainty estimate in i.i.d scenario. However, recent research has cast doubt about the validity of these uncertainty estimates under covariate shift \citep{ovadia2019can}. Moreover, the above methods do not make use of any unlabelled data for training or inference.

\paragraph{Semi-Supervised Learning}

Semi-supervised learning (SSL) covers the broad field of learning from both labelled and unlabelled data \cite{zhu2009introduction}. It s generally separated into two with most of the work covering \textit{inductive} SSL which aims to use the unlabelled data to learn a general mapping from the features to the outcome. Many recent works encourage the model to generalise better by using a \textit{regularisation} term computed on the unlabelled data \citet{berthelot2019mixmatch}. This includes \textit{entropy minimisation} which encourages the model to produce confident predictions on unlabelled data \cite{grandvalet2005semi,lee2013pseudo,jean2018semi} and \textit{consistency regularisation} which ensures the predictions for slightly perturbed data stay similar \cite{sajjadi2016regularization}. The other split covers \textit{transductive} SSL where the aim is to make predictions over only the unlabelled points given with no need to generalise further.
As we will show later, the proposed Transductive Dropout fits more into this framework, using the unlabelled data as a regulariser in order to induce a better variational approximation to the intractable posterior distribution.

However, our work is significantly different from traditional SSL in several ways.
First, we note that most existing works in SSL focus entirely on using unlabelled data to improve predictive performance (e.g. accuracy), but much less thoughts have been given to improving the uncertainty estimate for those predictions, which is the focus of this paper. 
Furthermore, our work explicitly addresses the issue of covariate shift between source and target data whereas traditional SSL often assumes that they are i.i.d.
In addition, most of the recent work in SSL considers problems like image classification and natural language processing where the methods can leverage the complicated dependencies in the features - we don't consider this a focus and develop a method that works appropriately for tabular data as well.

\paragraph{Unsupervised Domain Adaptation}

Unsupervised domain adaptation (UDA) is the task of training models to achieve better performance on a target domain, with access to only unlabelled data in the target domain and labelled data from a (different) source domain. \citet{kouw2019review} contains a detailed review of popular UDA methods. As with SSL, existing works on UDA centre around improving predictive performance rather than producing well-calibrated uncertainty estimates. Our work contributes to the UDA literature by proposing a method to improve the uncertainty estimates on the predictions made in the target domain.

\paragraph{Transfer Learning}
In the setting of transfer learning \cite{torrey2010transfer} the task does involve a change in distribution over features but typically also involves some amount of labels on the target set (known as one-shot or few-shot learning). This has led to a lot of work that uses the training set to learn a useful prior for a second model that can be trained on the labelled data in the target set \cite{raina2006constructing,karbalayghareh2018optimal}. Given the complete lack of labels in our target data set this is inapplicable for our problem.

\section{Preliminaries}
\subsection{Notation and Problem Setup}
Let $\mathbf{x} \in \mathbb{R}^d$ be a $d$-dimensional feature~vector,~and~$y \in \mathcal{Y}$ be the prediction target; where $\mathcal{Y} = \mathbb{R}$ for regression targets, and $\mathcal{Y} = \{1,\dots, K\}$ for $K$-class classification targets. We are presented with {\it two} sources of training data: a labelled data set $\mathcal{D}_L$, and an unlabelled data set $\mathcal{D}_U$. The labelled data set comprises a collection of $n$ feature-label pairs, i.e., $\mathcal{D}_L=\{(\mathbf{x}_i, y_i)\}^n_{i=1}$, whereas the unlabelled set comprises a collection of $m$ feature instances $\mathcal{D}_U=\{\mathbf{x}_j\}^m_{j=1}$.    

We assume that $\mathcal{D}_{L} = \{(\mathbf{x}_i,y_i)\}_{i=1}^n$ consists of i.i.d samples of features and labels drawn from the distribution 
\begin{align}
(\mathbf{x}_i,y_i) \sim p(\mathbf{x}) \times p(y|\mathbf{x}),\,\, \forall i \in \{1,\ldots,n\},\nonumber
\end{align} 
where both $p(\mathbf{x})$ and $p(y|\mathbf{x})$ are~unknown,~and~could~only~be accessed empirically through $\mathcal{D}_{L}$.~Throughout~the~paper,~we will refer to $p(\mathbf{x})$ as the {\it feature distribution} --- feature instances in the unlabelled data set are assumed to~be~drawn from a shifted feature distribution as follows:
\begin{align}
\mathbf{x}_j \sim p^{\prime}(\mathbf{x}),\,\, \forall j \in \{1,\ldots,m\},\nonumber
\end{align}
where $p^{\prime}(\mathbf{x}) \neq p(\mathbf{x})$, whereas the unobserved labels in the data set $\mathcal{D}_{U}$, i.e., the blue dots in Figure \ref{Fig2} corresponding to $\{y_j\}_{j=1}^{m}$, are generated from the same conditional distribution $y_j \sim p(y|\mathbf{x}_j)$. Note that even though the feature distributions $p'(\mathbf{x})$ and $ p(\mathbf{x})$ differ, the conditional $p(y|\mathbf{x})$ is invariant across the two data sets. This situation is commonly known as \textit{covariate shift} \cite{shimodaira2000improving}. We denote the entirety of observed data $\mathcal{D} = \{\mathcal{D}_L \cup \mathcal{D}_U \}$.

\subsection{Learning from (and for) unlabelled data}
Our key objective is to use the~(source)~labelled~data~set~$\mathcal{D}_{L}$ to train a model that would be applied to the (target) unlabelled data set $\mathcal{D}_{U}$. However, since the feature distributions in $\mathcal{D}_{L}$ and $\mathcal{D}_{U}$ mismatch, we cannot~expect~a~model~trained on $\mathcal{D}_{L}$ to perfectly generalise to $\mathcal{D}_{U}$. Thus, we aim at training the model to {\it learn} which prediction instances can be {\it confidently} transferred from $\mathcal{D}_{L}$ to $\mathcal{D}_{U}$, and which cannot be confidently generalised across the two distributions.~To~this end, we train the model to score its uncertainty on predictions issued for all feature instances in $\mathcal{D} = \{\mathcal{D}_L \cup \mathcal{D}_U \}$.    

We take a \textit{Bayesian} approach to uncertainty~estimation.~That is, for a model with parameter $\boldsymbol{\theta}$ and~a~test~point~$\mathbf{x}^* \sim p^\prime(\mathbf{x})$, the Bayesian posterior distribution over $y^*$ is 
\begin{equation}
\label{eq:bayesian}
\underbrace{p(y^* | \mathbf{x}^*, \mathcal{D})}_{\mbox{\footnotesize \bf Total uncertainty}} =  \int \underbrace{p(y^*| \mathbf{x}^*, \boldsymbol{\theta})}_{\substack{\mbox{\footnotesize \bf Data} \\ \mbox{\footnotesize \bf uncertainty}}} \underbrace{p(\boldsymbol{\theta} |\mathcal{D})}_{\substack{\mbox{\footnotesize \bf Model} \\ \mbox{\footnotesize \bf uncertainty}}} d\boldsymbol{\theta}. 
\end{equation}
The posterior decomposition in (\ref{eq:bayesian}) comprises two types of uncertainty \cite{malinin2018predictive}: {\it data uncertainty}, also referred to as aleatoric uncertainty,~is~the~variance~of~the~true conditional distribution $p(y|\mathbf{x})$, reflecting the inherent ambiguity or noise in the true labels $y$ \cite{gal2017concrete}.~The~second type of uncertainty, {\it model uncertainty}, pertains to the model's epistemic uncertainty created by the lack of training examples in the vicinity of the test feature $\mathbf{x}^*$. Since the conditional $p(y|\mathbf{x})$ is invariant across the source and target distributions, it is the model uncertainty that we focus on.

\subsection{Standard approximate Bayesian is not enough...}
A true Bayesian model (with appropriate priors) would completely capture model uncertainty in $\mathcal{D}_U$~by~simply~training the model on $\mathcal{D}_L$ in a supervised fashion, while completely ignoring the unlabelled data in $\mathcal{D}_U$ \cite{sugiyama2007mixture}. However, exact Bayesian inference in neural networks is generally~intractable~(and~computationally~expensive), hence existing practical solutions to Bayesian modelling rely on approximate inference schemes, for example based on Monte Carlo dropout (MCDP) \cite{gal2016dropout}.

While approximate inference via MCDP --- with appropriate hyper-parameter tuning --- provides reliable uncertainty estimates for in-distribution data (i.e., feature instances in $\mathcal{D}_L$), it has been shown in \citet{ovadia2019can} that these methods lead to miscalibrated estimates of uncertainty for out-of-distribution data. In the next Section, we develop an approximate Bayesian scheme that makes use of the unlabelled data in $\mathcal{D}_{U}$ to provide more accurate uncertainty estimates on the predictions made for features instances drawn from the shifted distribution $p^\prime(\mathbf{x})$.

\section{Transductive Regularisation}
How can we use our knowledge of the~unlabelled~data~in~$\mathcal{D}_{U}$ to improve the uncertainty estimates on predictions made for the target distribution $p^\prime(\mathbf{x})$? In this Section,~we~develop~an approximate Bayesian method tailored to this~task.~Here,~we regard a neural network (NN) as a distribution $p(y|\mathbf{x}, \mathbf{\theta})$ that assigns a probability to each possible output $y$.

\subsection{Variational inference with posterior regularisation}
In a Bayesian framework, we specify a prior distribution $p(\mathbf{\theta})$ on the NN parameters, and obtain the posterior $p(\mathbf{\theta}|\mathcal{D})$ via Bayes rule. In practice, the posteriors $p(\mathbf{\theta}|\mathcal{D})$ and $p(y|\mathbf{x}, \mathcal{D})$ in (\ref{eq:bayesian}) are both intractable. To address this issue, we use variational inference, whereby we use a surrogate distribution $q_\mathbf{\phi}(\mathbf{\theta})$ parameterised by $\phi$ to approximate $p(\mathbf{\theta}|\mathcal{D})$. The~parameter $\phi$ is obtained by minimising the KL-divergence between $p$ and $q$ as follows \cite{graves2011practical}:
\begin{equation}
\phi^* = \argmin_\phi \text{KL}\big[\,q_\phi(\mathbf{\theta})\,||\,p(\mathbf{\theta}|\mathcal{D})\,\big].
\label{varieq}
\end{equation}
In practice the KL divergence is not minimised directly, rather it is achieved my maximising the Evidence Lower BOund (ELBO), which can be written as:
\begin{equation}\label{eq:elbo}
    \mathcal{F}(\mathcal{D},\phi) = \mathbb{E}_{q_\phi} \big[ \log p(\mathcal{D} | \theta) \big] - \text{KL}\big[\,q_\phi(\mathbf{\theta})\,||\,p(\mathbf{\theta})\,\big],
\end{equation}
being seen as the balance of two terms. The objective being to maximise the log-likelihood under the surrogate distribution (first term) while regularising the approximation to not be too far from the prior (second term).
Variational inference also leads to an approximate posterior predictive distribution $q_\phi(y|\mathbf{x}, \mathcal{D})$, obtained by replacing $p(\mathbf{\theta}|\mathcal{D})$ in (\ref{eq:bayesian}) with its variational~counterpart~$q_\mathbf{\phi}(\mathbf{\theta})$.~Note that the unlabelled data in $\mathcal{D}_U$ is ancillary to the optimisation problem in (\ref{varieq}), since mere evidence maximisation would render $p(\mathbf{\theta}|\mathcal{D}_L)$ as the only relevant conditional for finding the variational parameter $\phi$. Hence, the vanilla variational Bayes is insufficient in our setup as it cannot capitalise on our knowledge of the unlabelled data in $\mathcal{D}_U$.    

To incorporate the unlabelled data in $\mathcal{D}_U$ into our inference machine, we resort to {\it posterior regularisation} \cite{zhu2014bayesian}. That is, instead of computing the variational posterior that best matches the true posterior in KL distance, we add a regulariser $\Omega$ to the objective in (\ref{varieq}), i.e.,
\begin{equation}\label{eq:rb}
\phi^* = \argmin_\phi \text{KL}\big[\,q_\phi(\mathbf{\theta})\,||\,p(\mathbf{\theta}|\mathcal{D})\,\big] + \Omega(q_\phi(\mathbf{\theta}|\mathcal{D})),
\end{equation}
in order to explicitly influence the learned variational posterior so that it produces the desired uncertainty profiles, i.e., posterior variance, over the target feature distribution $p^\prime(\mathbf{x})$. 

{\bf What do our sought-after uncertainty profiles look like?} 
In order to design the regulariser $\Omega$, we first need to specify the influences it needs to exert on the~learned~variational~posterior $q_\phi$. Let $\mathbb{E}[\,q\,]$ and $\mathbb{V}[\,q\,]$~denote~the~mean~and variance of a given distribution $q$, respectively. A ``good'' variational posterior is one that matches the true posterior $p(\mathbf{\theta}\,|\,\mathcal{D})$, and induces the following uncertainty profile:~for~any~pair~of~features $\mathbf{x}, \mathbf{x}^\prime \sim p^\prime(\mathbf{x})$ drawn from the target distribution, the variational posterior satisfies the following condition: 
\begin{align}
\mathbb{V}[\,q_\phi(y|\mathbf{x}, \mathcal{D})\,] \geq \mathbb{V}[\,q_\phi(y|\mathbf{x}^\prime, \mathcal{D})\,] \Leftrightarrow p(\mathbf{x}^\prime) \geq p(\mathbf{x}). 
\label{posterdesire}
\end{align}
That is, the variance of the variational posterior, which quantifies the model's uncertainty, should be smaller for target test points that are close (in distribution)~to~the~labelled~data in $\mathcal{D}_L$, and vice versa. The key idea behind~our~posterior~regularisation approach is that the augmentation of labelled and unlabelled data serve as ``pseudo-labels'' of model confidence --- by regarding the condition in (\ref{posterdesire}) as an auxiliary classification task wherein~$q_\phi$~predicts~whether~a~feature~$\mathbf{x}$~is drawn from the source or target distributions, we can ``train'' $q_\phi$ to make this binary prediction via its variance. Building on this insight, the rest of this Section builds a regulariser $\Omega$ that enables $q_\phi$ to discriminate source and target features.

\begin{figure*}[ht]
  \centering
  \includegraphics[width=6.5in]{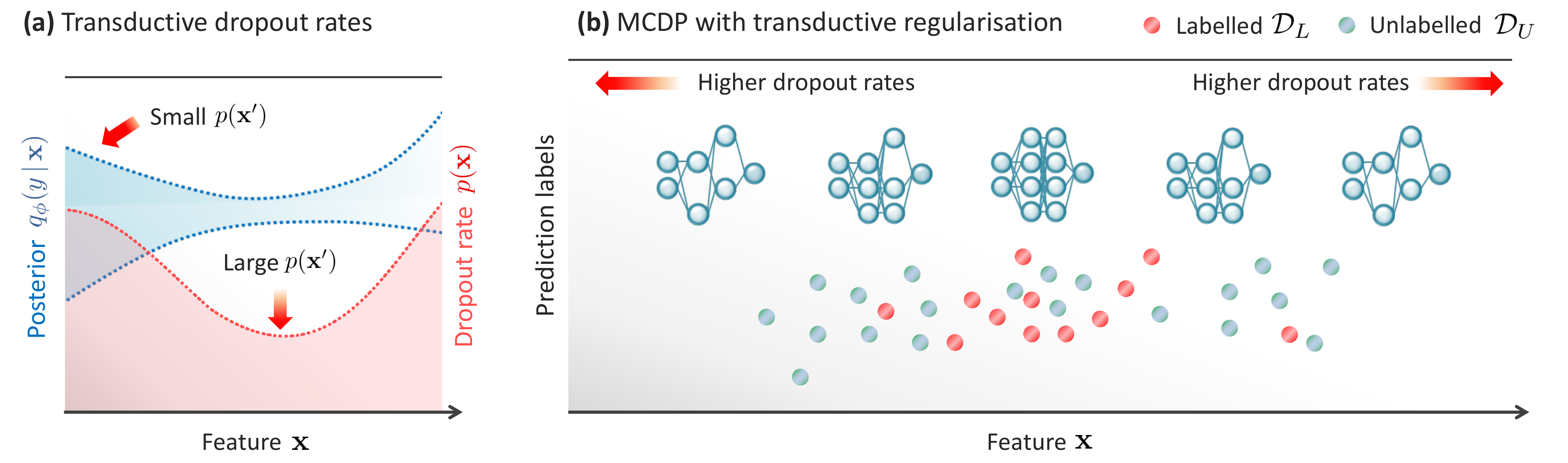}
  \caption{\footnotesize {\bf Pictorial depiction of transductive dropout inference.} (a) Here, we depict an exemplary one-dimensional feature space, along with the corresponding variational posterior $q_\phi(y|\mathbf{x})$ and feature-dependent dropout rate $p(\mathbf{x})$. Transductive dropout inference operates by adapting the dropout rate so that it induces larger posterior variance for regions with dense concentration of unlabelled data, but low density for labelled data (small $p(\mathbf{x}^\prime)$ for some $\mathbf{x}^\prime \sim p^\prime(\mathbf{x}))$. (b) This panel shows an exemplary realisation of labelled and unlabelled data sets for the same example in panel (a). Red dots are fully observed, whereas for blue ones, we only observe the locations but not the outputs on the $y$-axis. The typical behaviour of the transductive dropout is to increase the dropout rates in regions where unlabelled data are denser than labelled data, creating more variability in the Monte carlo samples of the network outputs. Here, exemplary instances of test-time dropout applied to the network architecture for different values of the feature $\mathbf{x}$ are depicted.} 
  \rule{\linewidth}{.75pt}   
  \label{Fig2} 
  \vspace{-7mm} 
\end{figure*}

\subsection{Posterior regularisation via transductive dropout}
As~discussed~above,~we~seek~a~variational~posterior~that~best fits the labelled data in $\mathcal{D}_L$, and discriminates~source~and~target data. Before proceeding, we first define an augmented data set $\widetilde{\mathcal{D}} = \{(\mathbf{x}_k, y_k, z_k)\}^{n+m}_{k=1}$, where
\begin{align}
  (\mathbf{x}_k, y_k, z_k) =
  \begin{cases}
   (\mathbf{x}_k, y_k, 0), & \text{$\forall\,\,(\mathbf{x}_k, y_k) \in \mathcal{D}_L$,}  \\
   (\mathbf{x}_{k-n+1}, *, 1), & \text{$\forall\,\,\mathbf{x}_{k-n+1} \in \mathcal{D}_U$,}  
  \end{cases} \nonumber  
\end{align}
where $*$ corresponds to a missing value for the label $y$. In addition, we define the monotonic function~$g: \mathbb{R}^+ \to [0,1]$~as a map from positive real values to the unit~interval.~Given~the~variational~distribution $q_\phi$, our prediction of whether~the~feature~$\mathbf{x}$ comes from the source or target distributions is 
\begin{equation}
\hat{z}_\phi(\mathbf{x}) \, \triangleq \, g\Big(\mathbb{V}[\,q_\phi(y|\mathbf{x}, \mathcal{D})\,]\Big),
\label{eqq5}
\end{equation}
which follows directly from the condition in (\ref{posterdesire}). Given (\ref{eqq5}), we define the regulariser $\Omega$ in (\ref{eq:rb}) as the cross-entropy loss between predicted and true auxiliary variables, $\hat{z}$ and $z$, i.e.,  
\begin{multline}
\Omega(q_\phi(\theta \,|\, \widetilde{\mathcal{D}})) \\ = \sum^n_{k =1} \log\big(1 - \hat{z}_\phi(\mathbf{x}_k)\big) + \sum^{n+m}_{k=n+1} \log\big(\hat{z}_\phi(\mathbf{x}_k)\big).
\end{multline}
Thus, our variational posterior is obtained by plugging the regulariser $\Omega (q_\phi(\theta \,|\, \widetilde{\mathcal{D}}))$ in (\ref{eq:rb}) and solving for $\phi$, with the optional inclusion of a hyperparamter $\lambda$ to control the level of regularisation.
The exact choice of $g$ can as well be controlled although from our experiments it made little difference, and we settled on $g(x)=1-\frac{1}{1+x}$.
We note that this regularisation scheme addresses the issue of over-confident predictions on the target set without taking the naive approach of just increasing the~variance~everywhere~---~it is balanced by the location of the source data set that will lower the variance in our appropriately confident locations. Since the regulariser above solves the {\it transductive} learning problem of classifying source and target data in a way that resembles semi-supervised learning \cite{rohrbach2013transfer}, we call $\Omega$ a transductive regulariser.~In~what~follows,~we propose a practical way to implement transductive regularisation within the MCDP approximate inference framework.

{\bf Transductive~Dropout.}~We~extend~the~MCDP approximate inference scheme in \cite{gal2016dropout} by applying our posterior regularisation penalty, and allowing the dropout rates to vary per data point, dependent on the feature values. By enabling the dropout rates~to~be~a~function~of $\mathbf{x}$, we provide more degrees-of-freedom to flexibly craft the posterior variance $\mathbb{V}[q_\phi]$ so that it accurately discriminates source and target data points. 

Let $p$ be the dropout rate of the~underlying~NN~model.~We~parameterise $p$ to be dependent on the feature value $\mathbf{x}$ as follows. Let $v_\beta(.)$ be a neural network with a sigmoid output layer and parameters $\beta$, i.e., $v_\beta:\mathbb{R}^d \to [0,1]$ maps feature values to dropout rates so that $p = v_\beta(\mathbf{x})$. This equates to an approximate posterior distribution over the NN weights:
\begin{align}
q_\phi(\mathbf{w}) = \prod^N_{i=1} (1-v_\beta(\mathbf{x}))^{\frac{w_i}{m_i}}\, v_\beta(\mathbf{x})^{\frac{m_i - w_i}{m_i}}
\end{align}
for $w_i \in \{m_i,0\}$, 0 otherwise, where $\mathbf{w} = \{w_i\}_i$ is the set of weights for the NN modelling the conditional distribution $q_{\phi}(y|\mathbf{x},\mathcal{D})$. This leaves an optimisation objective (of the form in (\ref{eq:rb})) over the variational parameters $\phi = \{\beta, \mathbf{m}\}$. Using the equivalence between KL minimisation and squared loss minimisation under dropout regularisation, we can write the objective function in (\ref{eq:rb}) as
\begin{align}
R(\phi) = \sum_{\mathbf{x}_i \in \mathcal{D}_L}\|\mathbb{E}[q_\phi(y_i\,|\,\mathbf{x}_i)]\|^2_2 + \Omega(q_\phi(\theta \,|\, \widetilde{\mathcal{D}})),
\label{objec}
\end{align}
with the possibility of adding an $L2$ regulariser $\|\phi\|^2_2$ as well. As we can see, this objective incorporates both labelled and unlabelled data: the data set $\mathcal{D}_L$ contributes to the first term, which is concerned with fitting the observed labels drawn from the source distribution, whereas the second term, which depends on the entire augmented data set $\widetilde{D}$, makes sure that the induced variational posterior is aware of the mismatch between source and target feature distributions. We can see that this scheme, as depicted in figure \ref{fig:overview}, acts in a similar way to (\ref{eq:elbo}), primarily optimising the likelihood of the data under the approximation while constrained by a requlariser on the form of the distribution, only now the regulariser induces more specific behaviour and makes use of $\mathcal{D}_U$.

The regulariser in (\ref{objec}) can be computed in backpropagation using sample estimates of the posterior variance as follows. Let $\tilde{\phi}$ be the current estimate of the variational parameters at a given iteration of the gradient descent~procedure.~To~evaluate the model loss and gradients, we use the MCDP forward pass to sample $M$ outputs $\{\hat{y}^1_k, \ldots, \hat{y}^M_k\}$ for every $\mathbf{x}_k$ in $\widetilde{\mathcal{D}}$, and compute a Monte Carlo sample estimate of the transductive regularisation term as follows:  
\begin{align}
\widehat{\Omega}(q_{\tilde{\phi}}(\theta \,|\, \widetilde{\mathcal{D}})) = g\Bigg(\frac{1}{M} \sum_{m=1}^M(\hat{y}^m_k-\bar{y}_k)^2 \Bigg).
\label{objec2}
\end{align}
Computations of the estimator in (\ref{objec2}) only involve the forward pass, and evaluating its gradients is straightforward.  

{\bf Key insights}~Figure~\ref{Fig2}~provides~a~pictorial~depiction~of~our transductive dropout inference procedure applied to an exemplary, one-dimensional feature space. A key insight is that transductive dropout inference learns to adapt the dropout rate so that it induces larger posterior uncertainty for regions with dense concentration of unlabelled data, but low density for labelled data.

\subsection{Limitations}

With the target data included in the training regime, it would seem that this method does not lend itself to online deployment where new test points come in over time. We note though that retraining is not practically necessary every time a new prediction needs to be made, indeed given an initial collection of test points it would only be useful when we encounter a significant amount of new data that is covariate shifted further even than the original targets.

\section{Experiments}

\subsection{Toy Dataset}

\begin{figure*}
\centering
\includegraphics[width=1.0\textwidth]{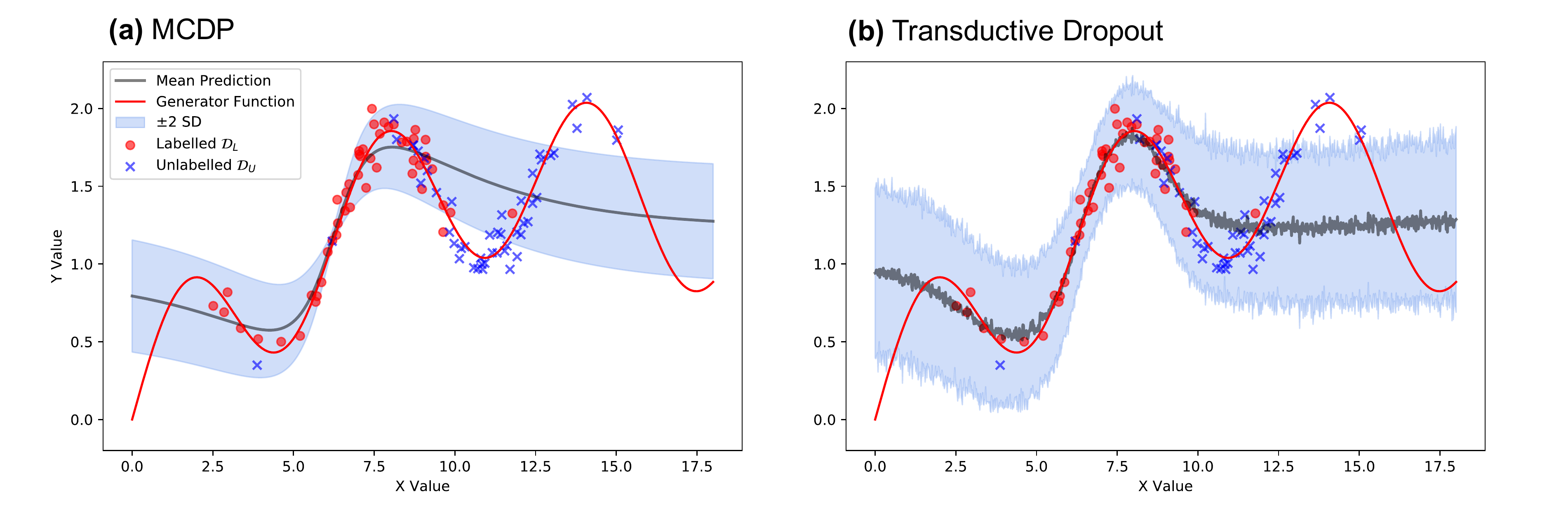}
\vspace{-7mm}
\caption{\footnotesize \textbf{Comparison of uncertainty predictions} (a) Here we show the $95\%$ confidence intervals for MCDP, demonstrating that while appropriate over the labelled data, they remain overconfident at the unlabelled data. (b) This panel shows the predictions for transductive dropout - the mean prediction remaining equally as accurate while producing more uncertainty at the unlabelled locations}
\label{fig:toy_nn} 
\rule{\linewidth}{.75pt} 
\vspace{-7mm} 
\end{figure*}

\begin{figure}[ht]
\centering
\includegraphics[width=1.0\columnwidth]{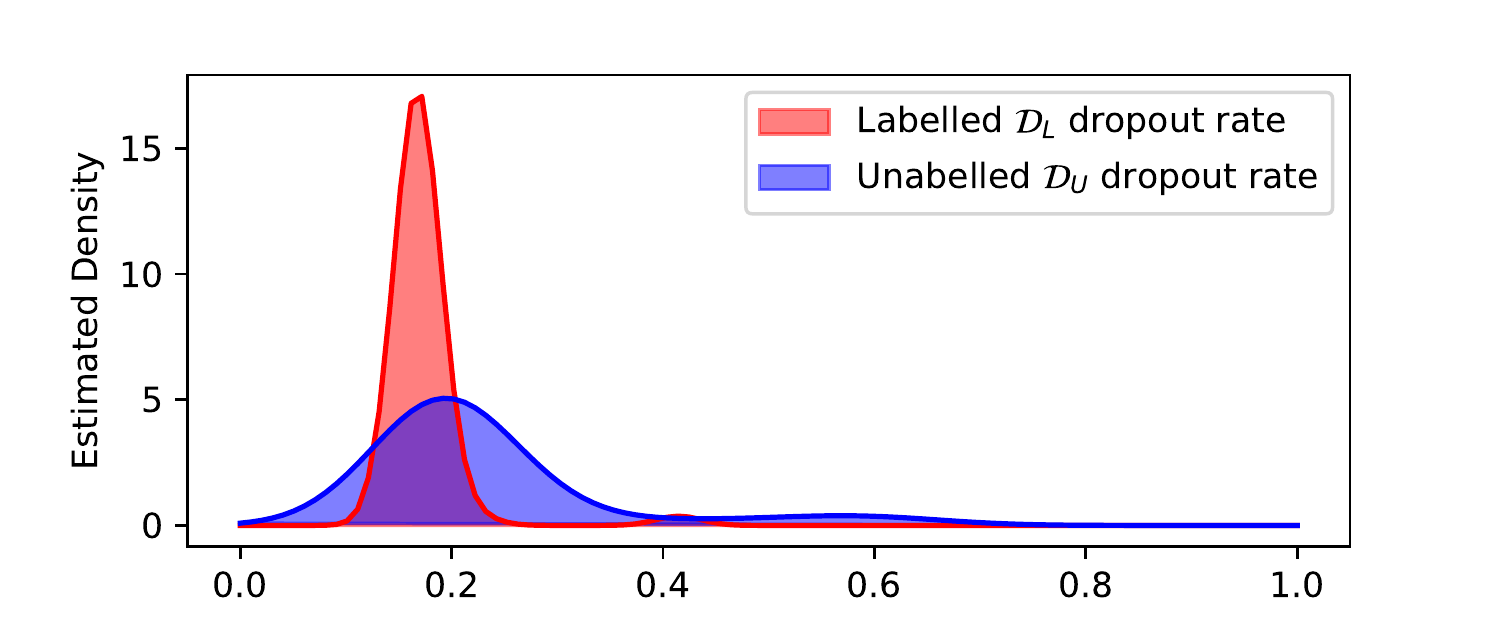}
\vspace{-7mm}
\caption{\footnotesize \textbf{Dropout rate distribution} Smoothed density estimates of the dropout rate distribution over the source and target sets.}
\label{fig:drop_rates} 
\rule{\linewidth}{.75pt} 
\vspace{-7mm} 
\end{figure}

In this section, we consider a toy 1-d regression example to show how standard BNNs produce overconfident uncertainty estimates under covariate-shift. 

The features in the source and target data sets are generated i.i.d. from $p(x) = N(7,4)$ and $p'(x) = N(11,4)$ respectively. The labels are generated as $y = f(x) + \epsilon$ where $f(x) = \frac{\sin{x}}{2} + \frac{x}{4} + \frac{x^2}{100}$ and $\epsilon$ is zero-centred Gaussian noise with standard deviation $0.1$. The source and target data sets contain 50 data points each.

Figure \ref{fig:toy_nn} compares the fit of MCDP (left) and transductive dropout (right). Both networks consist of two fully connected hidden layers with 32 and 64 neurons per respective layer with tanh activations. We see they produce similar mean predictions near the labelled training points. However, MCDP starts to issue over-confident predictions as the feature distribution shifts away from the training data. On the other hand, transductive dropout learns to output larger uncertainty in the areas of low density under $p(x)$. In figure \ref{fig:drop_rates} we plot smoothed density estimates of the learnt dropout rates for both source and target distribution points. In the source distribution the learnt rates are all quite tight around 0.18, while in the target distribution there is a much bigger spread reflective of the points' distances from labelled data.

\subsection{Prostate Cancer Mortality Prediction}

\paragraph{Background} Prostate cancer is the third most common cancer in men, with half a million new cases each year around the world \citep{quinn2002patterns}. It is far more common among the elderly with around 75\% of cases occur in men aged over 65 years.  Therefore, prostate cancer is  expected to bring increasing healthcare burden to countries with ageing population \citep{hsing2000international}. The latest clinical guideline for prostate cancer treatment recommends watchful waiting or non-invasive treatment for early-stage patients who have \textit{low mortality rate} \citep{heidenreich2011eau}. Surgery (Radical Prostatectomy) is recommended instead for high-risk patients whose health condition deteriorates rapidly. The patient's survival outlook therefore plays an important role in the treatment decisions. Hence, improved accuracy and uncertainty quantification for mortality prediction will help clinicians to design effective treatment plans and improve patients' life expectancy. 

\paragraph{Dataset}

We consider the problem of predicting and estimating the uncertainty of the mortality rate for patients with prostate cancer. 
Our training data consists of 240,486 patients enrolled in the American SEER program \citep{SEER}, while for our target data we consider a group of 10,086 patients enrolled in the British Prostate Cancer UK program \citep{UK}.
For both sets of patients we have identical covariate data with information concerning the age, PSA, and Gleason scores as well as what clinical stage they're at and which, if any, treatment they are receiving.
Note that while we have the same features for both sets this is an area where we expect a level of covariate shift given the different programs and the transition from American to British patients. Indeed we do see this, without giving a full break down of the summary statistics, patients in the Prostate Cancer UK are in general older with higher Gleason scores though not as far along in the clinical stages.

\paragraph{Benchmarks} We compare our method against competitive methods from the probabilistic deep learning literature based on their prevalence and applicability. While we consider this work quite different to semi-supervised learning, which do not usually consider improving uncertainty estimates,  we also include MixMatch as a benchmark \citep{berthelot2019mixmatch}. The methods we consider are:
\begin{enumerate}[noitemsep]
    \item \textit{MLP} - Standard feed forward neural network to benchmark accuracy.
    \item \textit{Dropout} - Monte Carlo dropout with rate 0.5 \cite{gal2016dropout,srivastava2014dropout} 
    \item \textit{Concrete Dropout} - Dropout with the rate treated as an additional variational parameter and is optimised with respect to the ELBO \cite{gal2017concrete}.
    \item \textit{Ensemble} - Ensemble of feed forward MLPs \cite{lakshminarayanan2017simple} with $K=10$ the number of models in the ensemble.
    \item \textit{MixMatch} - We implement a version of the MixMatch algorithm \cite{berthelot2019mixmatch} where we perform one round of label guessing and mixup and without sharpening. As the base predictive model we use a MC Dropout network.
    \item \textit{Last Layer Approximations (LL)} - Approximate inference for only the parameters of the last layer of the network \cite{riquelme2018deep}, using Dropout.
    \item \textit{Transductive Dropout - No Regularisation (TDNR)} - We implement transductive dropout as described above but without the addition of our variance regulariser to show that the gains are not just down to the ability to adapt the dropout rate to the input.
\end{enumerate}

For all of the neural networks we consider the same architecture of two fully connected hidden layers of 128 units each and tanh activation function. The initial weights are randomly drawn from N(0, 0.1) and all networks are trained using Adam \citep{kingma2014adam}. Hyperparameter optimisation remains an open problem under covariate shift - we used a validation set consisting of 10\% of the labelled data selected, not entirely randomly, but based on propensity score matching in order to obtain a set more reflective of the target data. With this, hyperparemeters were selected for all model through grid search.

\begin{table*}[ht]
\caption{Area under the ROC curve for two tasks, first correctly predicting the mortality rate of patients in the test set and secondly predicting whether for a given patient the model will make an error. We also report the average confidence interval (CI) length over test predictions, the average standard deviation (SD) at miss-classified points, and the increased number of patients receiving treatment (INPT) using the associated uncertainty in the model and a risk level of $15\%$.}
\label{table:mort_auroc}
\vskip 0.15in
\begin{center}
\begin{small}
\begin{sc}
\begin{tabular}{lccccccr}
\toprule
Method & Test Perf. & Error Pred. & CI Width & Misclassified SD & INPT \\
\midrule
\midrule
MLP   & 0.720 $\pm$ 0.012 & N/A & N/A & N/A & N/A \\
\midrule
MC Dropout      & 0.729 $\pm$ 0.016 & 0.730 $\pm$ 0.016 & 0.093 & 0.025 & 8\\
\midrule
Concrete Dropout   & 0.791 $\pm$ 0.012 & 0.794 $\pm$ 0.012 & 0.151 & 0.066 & 76\\
\midrule
Ensemble      & 0.761 $\pm$ 0.014 & 0.782 $\pm$ 0.014 & 0.037 & 0.018 & 8\\
\midrule
MixMatch   & 0.728 $\pm$ 0.016 & 0.726 $\pm$ 0.016 & 0.082 & 0.021 & 0\\
\midrule
LL     & 0.723 $\pm$ 0.014 & 0.696 $\pm$ 0.014 & 0.073 & 0.028 & 22\\
\midrule
TDNR   & 0.836 $\pm$ 0.010 & 0.808 $\pm$ 0.011 & 0.197 & 0.068 & 18\\
\midrule 
Transductive Dropout & 0.861 $\pm$ 0.009 & 0.857 $\pm$ 0.009  & 0.130 & 0.110 & 189\\

\bottomrule

\end{tabular}
\end{sc}
\end{small}
\end{center}
\vskip -0.1in
\end{table*}

\paragraph{Evaluation metrics} We consider five evaluation metrics for a comprehensive understanding of the model performance. First, we consider the prediction accuracy as measured by AUROC shown as ``TEST PERF.'' in table \ref{table:mort_auroc} \citep{bewick2004statistics}.  
Second, we consider the standard deviation of the posterior predictive distribution as a (unnormalised) predictor for whether or not the model will make an error on a given input. The corresponding AUROC score (``ERROR PRED'') measures the agreement between model uncertainty and the chance to predict wrongly, and hence reflects whether the model is well-calibrated.
Third, we present the average width of the 95\% predictive interval as a measure of general model confidence on unlabelled data (``CI WIDTH''). 
Next, we show the standard deviation of the predictive distribution on misclassified data (``MISCLASSIFIED SD''). 
Finally, we show the increased number of patients receiving treatment (INPT) using the associated uncertainty in the model and a risk level of $15\%$. All quantities related to the posterior distribution are estimated by MC sampling.

\paragraph{Main results} First, we note that transductive dropout yields an improvement in the  AUROC on the mortality prediction against the other benchmarks, demonstrating that our improved uncertainty calibration does not come at the cost of mean accuracy.
Our focus though is on the calibration of our uncertainty estimates. While ultimately it is impossible to properly test how close uncertainty predictions are to what would be the \textit{true} uncertainty, we test by using the posterior predictive variance to classify whether or not the model will make a mistake. The intuition here is that if the model is appropriately uncertain the variance will be high when a mistake is likely and low when not, thus a high performance on using variance as a predictor for when the model will make a mistake should demonstrate appropriate uncertainty estimates. Here we see that transductive dropout significantly outperforms the other benchmarks, suggesting that in general the high variance predictions are indeed associated with those that are more likely to be wrong. We additionally focus on these predictions that each method gets wrong and look at the average standard deviation at each of these points. Here transductive dropout shows on average it's much less confident about its incorrect predictions than the the other benchmarks, which is the preferred behaviour. It is important to note that this is not at the expense of confidence over all predictions as we show that both concrete dropout and TDNR both have on average larger confidence intervals than transductive dropout.

\begin{figure}[ht]
\centering
\includegraphics[width=1.0\columnwidth]{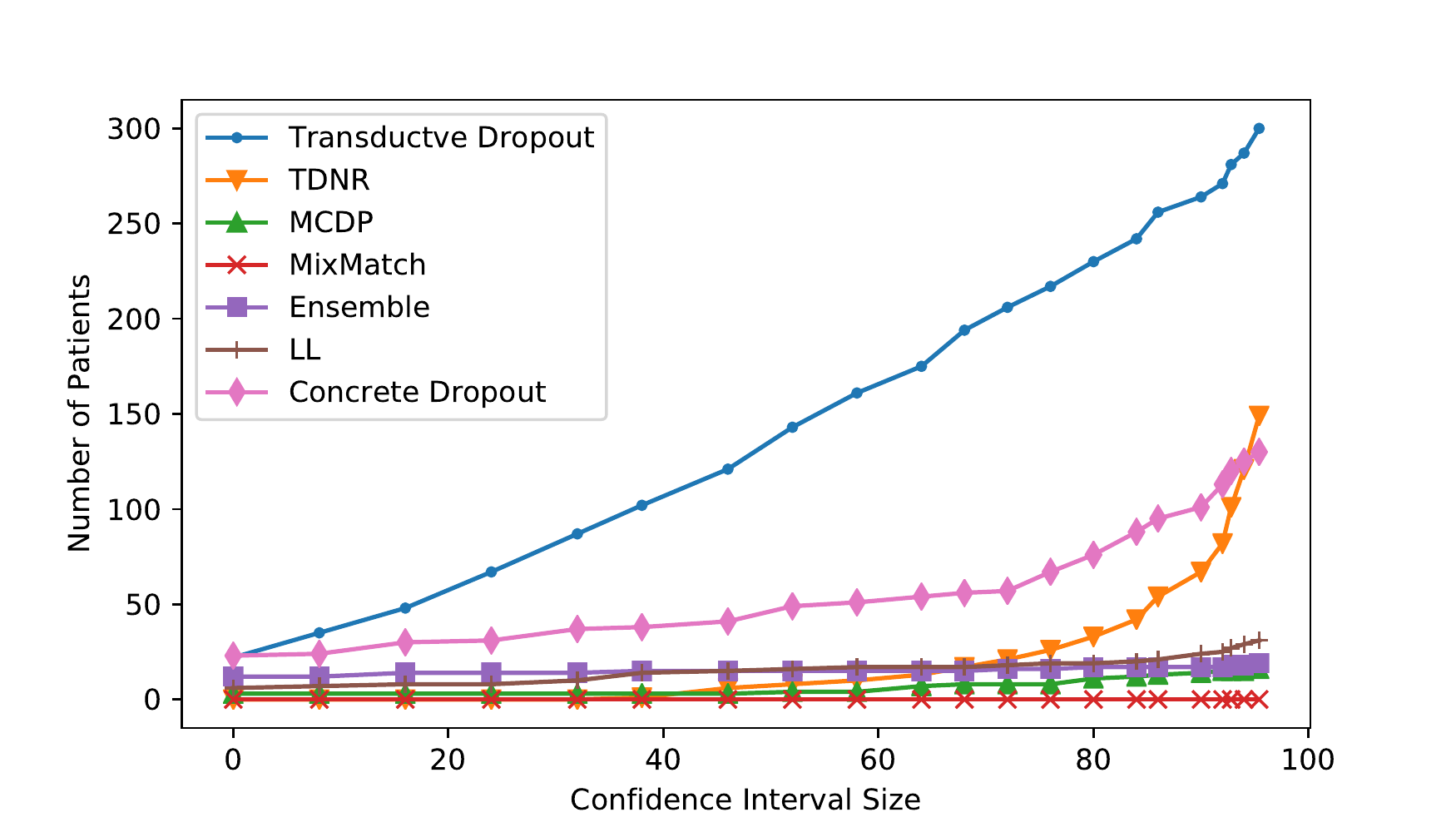}
\vspace{-9mm}
\caption{\footnotesize \textbf{Improved patient outcome} We show how many more patients, for a risk threshold of $50\%$, correctly receive treatment as the size of the confidence interval on the prediction of risk changes.}
\label{fig:conf} 
\rule{\linewidth}{.75pt} 
\vspace{-7mm} 
\end{figure}

\paragraph{Impact on patients} Given our motivations we also ground the performance of our method in how it could be used in real world decision making on the treatments offered to patients. There are many reasons treatment options may not be offered to patients including cost and potential side effects, as such there will usually be an associated risk level which a patient must be above in order to receive treatment. It's thus very damaging to patients for a model to confidently predict them to be low risk when they are indeed not. In Table \ref{table:mort_auroc} we set a $15\%$ threshold, and show how many more patient would receive treatment if we consider coverage of the $95\%$ confidence interval on the patients risk, with the assumption that these cases can be handed off to a human expert who will correctly classify them. We see that transductive dropout results in a large increase in previously patients misclassified as low risk receiving treatment and we develop the impact on treatment options further in Figure \ref{fig:conf}. Here we set a treatment risk threshold at $50\%$ and show how the size of any predicted confidence intervals over a patients risk impacts the increased number of patients correctly receiving treatment. Naturally for all methods as the confidence interval grows the number of now correctly treated patients increases but transductive dropout consistently outperforms the other benchmarks as it is less often confidently incorrect in its risk prediction.

\textbf{How does the covariate-shift affect uncertainty?} 
Of interest is to consider how the covariate shift has actually impacted our models performance. To that end we consider the feature distribution of those points misclassified by the model to see how it compares to both source and target sets. 
One of the most important factors affecting both the treatment and survival of prostate cancer patient is the age at diagnosis \citep{bechis2011impact}. Studies have shown that older patients tend to have worse survival outlook and are more likely to receive surgery \citep{hall2005impact}. 
In our source data, the average age at diagnosis is 66 years old moving up to 70 in the target set. Comparing to the distribution over ages for incorrectly predicted cases, where the average is 74, we see that it is for the patients who are considerably older than those usually seen in the training data that the model is less sure about. 
We see a similar story in their PSA scores (measurements of \textit{prostate specific antigen} in the patients blood). 
PSA score is known to be a highly sensitive indicator for the risk level and severity of prostate cancer, and it is widely adopted in cancer screening and monitoring \citep{grimm2012comparative}. 
Again we see an increase in the average from 14.8 to 18.4 from source to target set but for those that are incorrectly classified the mean is much higher at 28.6.§ 
The percentage of patients receiving surgery in the incorrectly classified group is twice that of those correctly classified, suggesting that our models are least confident in areas which we might think are the most at risk given domain knowledge - the more elderly with high levels of PSA. The model struggles with them (is much less confident) though as they are values which don't have high density in the training data, demonstrating that blind application of a model to a covariate shifted data set may easily yield surprisingly incorrect predictions. Fortunately transductive dropout tends to return high uncertainty over its predictions on this covariate shifted data such that the practitioner can suitably inform any decisions to be taken as a result of these predictions.

\subsection{Additional Results}

We focused here on an important real-world example problem, but with the aim of demonstrating generalisation,
we provide further benchmark results on some publicly available data sets from the UCI machine learning repository \citep{Dua:2019} in appendix A. 

\section{Conclusions}

In this paper we introduced transductive regularisation, a method for using unlabelled data to calibrate the variance of Bayesian neural networks by introducing the auxiliary task of using the posterior predicted variance to discriminate between source and target distributions. We showed that this amounts to performing posterior regularisation in approximate Bayesian inference and results in more useful uncertainty predictions. We examined an instantiation of this framework within MCDP, transductive dropout, and demonstrate its applicability in the real task of predicting prostate cancer mortality, where it outperforms the tested benchmarks and demonstrates a higher level of appropriate uncertainty calibration.

\paragraph{Future Work}
The question of perfect calibration is clearly not solved and there are several immediate directions for further work that present themselves. First is an extension to frequentist probabilistic ensembles, this is not entirely trivial since the capturing of model uncertainty comes from averaging across elements of the ensemble making estimates during training more complicated to obtain. Second would be an application in the active learning setting and using targeted labelled data acquisition in order to appropriately reduce the model uncertainty - the sub-network used to predict the rate in transductive dropout could inform an important part of an acquisition function.

\section*{Acknowledgements}

We would like to thank the anonymous reviewers for their helpful comments and suggestions. Research in this paper was supported by the National Science Foundation (NSF grants 1524417 and 1722516), and the US Office of Naval Research (ONR).

\bibliography{example_paper}
\bibliographystyle{icml2020}

\onecolumn
\appendix
\section{Additional Results}
\label{app:add_res}

We provide some additional results in table \ref{table:add_res} on publicly available data sets taken from the UCI machine learning repository. Specifically we take three data sets: \textit{Breast Cancer}, \textit{Iris}, and \textit{Wine} before slightly adapting them to more naturally fit the covariate shifted setting. First we make them a binary classification problem by taking the class with the largest members as positive and all others as negative. We then split the data into training and testing sets by projecting on to the first principal component and sampling a 20\% testing set weighted by this value. 

For all of the neural networks we consider the same architecture of two fully connected hidden layers of 32 and 64 hidden units each with tanh activation function. The initial weights are randomly drawn from N(0, 0.1) and all networks are trained using Adam. We consider the prediction accuracy as measured by AUROC shown as ``TEST PERF.'' as well as the standard deviation of the posterior predictive distribution as a (unnormalised) predictor for whether or not the model will make an error on a given input. The corresponding AUROC score (``ERROR PRED'') measures the agreement between model uncertainty and the chance to predict wrongly, and hence reflects whether the model is well-calibrated.

We see tranductive dropout always performs strongly on test performance, and though not always the best is certainly competitive in all cases, demonstrating their doesn't appear to be a toll on mean predicitive power. Further though we see that transductive dropout does remain the best across the data sets on the task of error prediction, demonstrating better uncertainty calibration, the focus of this work.

\begin{table*}[ht]
\caption{For the three datasets we present the area under the ROC curve for two tasks, first correctly predicting the classification in the test set and secondly predicting whether for a given test point the model will make an error.}
\label{table:add_res}
\vskip 0.15in
\begin{center}
\begin{small}
\begin{sc}
\begin{tabular}{lcccccr}
\toprule
Method & \multicolumn{2}{c}{Breast Cancer} & \multicolumn{2}{c}{Iris} & \multicolumn{2}{c}{Wine} \\
& Test Perf. & Error Pred. & Test Perf. & Error Pred. & Test Perf. & Error Pred. \\
\midrule
\midrule
MC Dropout      & 0.979 $\pm$ 0.012 & 0.662 $\pm$ 0.033 & 0.937 $\pm$ 0.044 & 0.063 $\pm$ 0.046 & 0.972 $\pm$ 0.026 & 0.775 $\pm$ 0.155\\
\midrule
Concrete \\ Dropout   & 0.791 $\pm$ 0.006 & 0.794 $\pm$ 0.006 & 0.952 $\pm$ 0.038 & 0.847 $\pm$ 0.055 & 1.000 $\pm$ 0.000 & 0.915 $\pm$ 0.050\\
\midrule
Ensemble      & 0.978 $\pm$ 0.011 & 0.675 $\pm$ 0.007 & 0.960 $\pm$ 0.041 & 0.571 $\pm$ 0.115 & 0.993 $\pm$ 0.007 & 0.939 $\pm$ 0.037\\
\midrule
MixMatch   & 0.986 $\pm$ 0.010 & 0.529 $\pm$ 0.046 & 0.242 $\pm$ 0.069 & 0.758 $\pm$ 0.069 & 0.889 $\pm$ 0.050 & 0.611 $\pm$ 0.105\\
\midrule
LL     & 0.950 $\pm$ 0.033 & 0.329 $\pm$ 0.032 & 0.929 $\pm$ 0.064 & 0.071 $\pm$ 0.064 & 0.986 $\pm$ 0.013 & 0.575 $\pm$ 0.230\\
\midrule
TDNR   & 0.979 $\pm$ 0.013 & 0.945 $\pm$ 0.026 & 0.940 $\pm$ 0.045 & 0.657 $\pm$ 0.105 & 1.000 $\pm$ 0.000 & 0.890 $\pm$ 0.055\\
\midrule 
Transductive \\ Dropout & 0.968 $\pm$ 0.017 & 0.975 $\pm$ 0.015  & 0.956 $\pm$ 0.045 & 0.877 $\pm$ 0.082 & 1.000 $\pm$ 0.000 & 0.951 $\pm$ 0.034\\

\bottomrule

\end{tabular}
\end{sc}
\end{small}
\end{center}
\vskip -0.1in
\end{table*}

\end{document}